\documentclass[letterpaper]{article} 
\usepackage{aaai2027}  
\usepackage[hyphens]{url}  
\usepackage{graphicx} 
\usepackage{natbib}  
\usepackage{caption} 
\usepackage{algorithm}
\usepackage{algorithmic}

\usepackage{newfloat}
\usepackage{listings}
\DeclareCaptionStyle{ruled}{labelfont=normalfont,labelsep=colon,strut=off} 
\floatstyle{ruled}
\newfloat{listing}{tb}{lst}{}
\floatname{listing}{Listing}

\usepackage{booktabs}
\usepackage{multirow}
\usepackage{makecell}
\usepackage{enumitem}
\usepackage{amsmath}
\usepackage{amssymb}
\usepackage{xcolor}
\definecolor{myyellow}{cmyk}{0,0.4,1,0.15}
\definecolor{myblue}{cmyk}{1,0.6,0,0.15}
\definecolor{mygreen}{cmyk}{1,0,0.9,0.15}

\newcommand{\myparagraph}[1]{\vspace{0.05\baselineskip}\noindent{\textbf{#1.}}~}
\title{NeuCME: Toward Dynamic Multimodal Continual Learning via Neural Combinatorics of Multiple Experts}
\author{
    Kai Guo\textsuperscript{\rm 1},
    Chuanbin Liu\textsuperscript{\rm 4},
    Peng Hu\textsuperscript{\rm 2},
    Hao Wang\textsuperscript{\rm 1}\corresponding,
    Xi Peng\textsuperscript{\rm 2, \rm 3}\\
}
\affiliations{
    \textsuperscript{\rm 1}College of Computer Science, Sichuan University, Chengdu, China\\
    \textsuperscript{\rm 2}School of Artificial Intelligence, Sichuan University, Chengdu, China\\
    \textsuperscript{\rm 3}Tianfu Jincheng Laboratory, Chengdu, China\\
    \textsuperscript{\rm 4}School of Economics and Management, China University of Petroleum (Beijing), Beijing, China\\

    \{kaiguo.gm, penghu.ml, pengxi.gm, cshaowang\}@gmail.com, liuchuanbingl@163.com
}

\begin{document}
\nocopyright
\maketitle

\begin{abstract}
Multimodal continual learning has recently shown great potential for developing agents with human-like intelligence by continuously learning new tasks across multiple modalities. However, existing methods typically assume that the set of modalities per task is predefined and fixed. In this paper, we investigate a more realistic learning setting, referred to as dynamic multimodal continual learning, in which the set of modalities may vary across tasks rather than remaining fixed. This setting involves two primary challenges: (i) spatio-temporal catastrophic forgetting and (ii) adaptive multimodal fusion. To address these challenges, we propose NeuCME (as shorthand for \textbf{Neu}ral \textbf{C}ombinatorics of \textbf{M}ultiple \textbf{E}xperts), a novel framework designed to effectively learn and integrate knowledge across tasks with varying modalities. The proposed NeuCME model comprises three key components, namely modality-combinational rehearsal, multi-gated mixture-of-experts, and task relevance-guided distillation. Furthermore, we formulate an evaluation metric to quantify the dynamism of task sequences and then set up a comprehensive benchmark with different degrees of dynamism. Extensive experiments using four real-world datasets demonstrate that the proposed NeuCME outperforms state-of-the-art methods markedly.~\footnote{Code will be released upon acceptance.}
\end{abstract}


\section{Introduction}
\label{sec:intro}

With the utilities of powerful multimodal models that integrate images, text, video, and audio, multimodal continual learning (MCL) has emerged as a pivotal paradigm in modern AI research \cite{liu2023ai,yu2024recent}. In a nutshell, MCL aims to enable models to continuously learn from multimodal data streams while accumulating previously acquired knowledge and adapting to new tasks without retraining from scratch \cite{kim2024multi,sarfraz2025beyond}.
Despite its potential to foster human-like intelligence, most existing MCL methods are limited to bimodal tasks (typically vision and text \cite{zheng2023prevent,yu2024boosting}) and assume a fixed multimodal setting where all tasks involve the same set of modalities \cite{liu2025c}, restricting their real-world applicability. In real-world applications such as robotics, conversational agents, and autonomous driving \cite{LadakWLA25}, the operating environment is inherently open and dynamic. For instance, certain modalities may be lost over time due to sensor failures, while new modalities may emerge as required by the downstream application. In other words, the set of modalities may vary across tasks rather than remain fixed. In this paper, we focus on this open and dynamic learning scenario, referred to as \textit{dynamic multimodal continual learning} (dynamic MCL). Unlike conventional settings of MCL, dynamic MCL not only considers the modality-inconsistent \cite{pian2024modality} or modality-incremental~\cite{yu2024llms} scenario but also encompasses the missing-modality learning problem~\cite{guo2025efficient}.

\begin{figure}
    \centering
    \includegraphics[width=\linewidth]{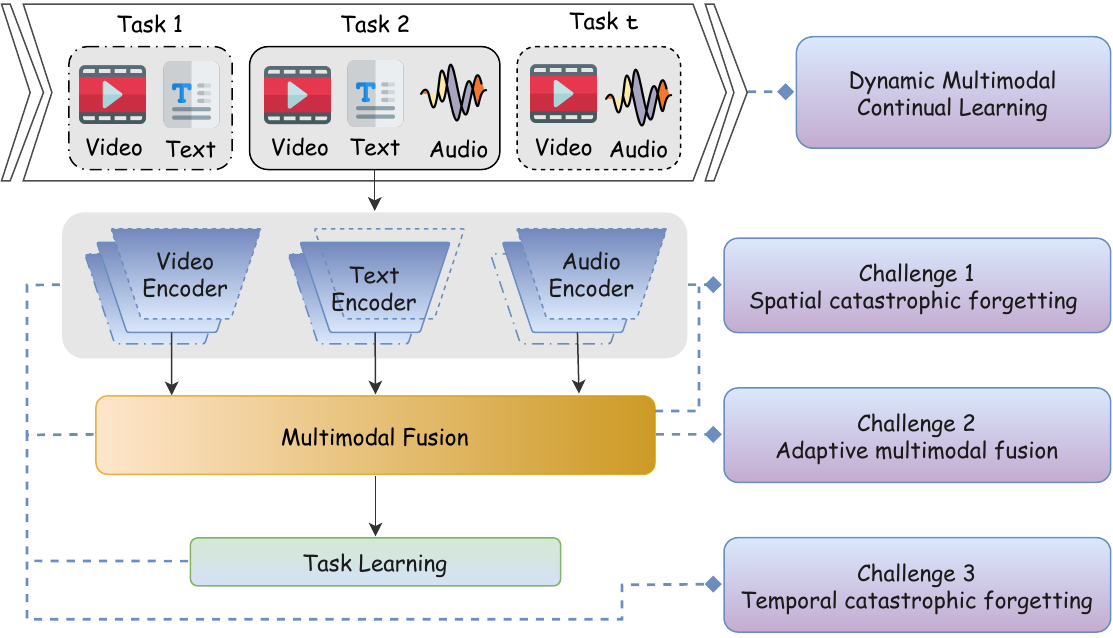}
    \caption{An illustrative example of our motivation. \textit{Left}: A schematic of dynamic MCL that learns a sequence of tasks, where Task 1 includes video and text, Task 2 has video, text and audio, and Task $t$ contains video and audio modalities. 
    \textit{Right}: Illustration of three fundamental challenges involved in modality encoding, multimodal fusion, and task learning.}
    \label{fig:challenges}
\end{figure}
Figure \ref{fig:challenges} illustrates a schematic of dynamic MCL and its fundamental challenges. As shown in this figure, a dynamic MCL task sequence could have three types of modality changes among adjacent tasks. Specifically, Task 2 includes a new modality (i.e., audio) than Task 1. Task $t$ contains only video and audio rather than the three modalities as in Task 2. Also in this task sequence, Task $t$ involves video and audio, but Task 1 involves video and text. Given these cases, we generalize such heterogeneous tasks into three types: i) \textit{Modality-augmented task}, referred to as a task in which one or more new modalities are added compared to previous tasks (e.g., Task 2 vs. Task 1). ii) \textit{Modality-reduced task}, referred to as a task in which one or more modalities are removed compared to previous tasks (e.g., Task $t$ vs. Task 2). iii) \textit{Modality-mixed task}, referred to as a task in which modality augmentation and modality reduction occur simultaneously compared to previous tasks (e.g., Task $t$ vs. Task 1).

Unlike conventional MCL, in which all tasks contain the same modalities, dynamic MCL needs to handle the aforementioned heterogeneous tasks, where each task in a given sequence may involve different modalities. This setting raises several new challenges, including \textit{spatio-temporal catastrophic forgetting} and \textit{adaptive multimodal fusion}. We elaborate on each of these challenges below.

As discussed in \citet{yu2024recent}, MCL inherently faces a core challenge, namely \textit{catastrophic forgetting} (CF). In the dynamic MCL setting, CF further exhibits two distinct forms, which we refer to as (1) \textit{spatial CF} and (2) \textit{temporal CF}. Specifically, at any point in time, suppose a model has learned $t$ tasks, and now the model heads for a new task $T_{t+1}$. The model would suffer from spatial CF if task $T_{t+1}$ falls into one of the three aforementioned task types, i.e., modality-augmented, modality-reduced, or modality-mixed. For example, when a previously learned modality is absent in the new task, the model may forget knowledge associated with that modality after the new task is learned \cite{guo2025efficient}. Besides, incorporating new modalities may negatively interfere with the retention of knowledge from previously learned modalities \cite{song2025harmony}. Since this form of forgetting arises from changes in the modality space, we call it \textit{spatial CF}. On the other hand, temporal CF arises as the model adapts to new tasks over time, during which previously acquired knowledge tends to be overwritten or gradually degraded, resulting in noticeable performance drops on previously encountered tasks \cite{kirkpatrick2017overcoming}. Since this form of forgetting unfolds over time, we call it \textit{temporal CF}.  In dynamic MCL, temporal CF becomes even more intricate, as continual changes in the modality space further intensify interference across tasks. The interaction between temporal task evolution and spatial modality dynamics ultimately amplifies the forgetting effect, making it challenging to retain long-term knowledge. 



Last, in dynamic MCL settings, input data per task may arrive with a different set of modalities, yet these heterogeneous modalities should still be effectively fused into a unified representation for downstream applications. However, existing MCL methods are typically designed under modality-fixed assumptions, relying on predetermined fusion schemes, such as concatenation or attention-based methods \cite{cai2023task}. These modality-fixed approaches fundamentally fail to generalize to dynamic modality scenarios \cite{baltruvsaitis2018multimodal}, where the model needs to dynamically adjust its structure on the fly to maintain effective multimodal fusion. Thus, the ability of \textit{adaptive multimodal fusion} that can dynamically activate relevant components for any given set of modalities is essential and nontrivial for building reusable MCL systems.

To address the three fundamental challenges, i.e., spatial CF, temporal CF, and adaptive multimodal fusion, this paper presents \underline{\textbf{Neu}}ral \underline{\textbf{C}}ombinatorics of \underline{\textbf{M}}ultiple \underline{\textbf{E}}xperts (\textbf{NeuCME}). The NeuCME model exhibits an amalgamation of \textit{\textbf{modality-combinational rehearsal}} (Section 3.3), \textit{\textbf{multi-gated mixture-of-experts}} (Section 3.4), and \textit{\textbf{task relevance-guided distillation}} (Section 3.5) to simultaneously preserve past knowledge, adaptively fuse heterogeneous modalities, and effectively accommodate new tasks in dynamic environments with multiple degrees of dynamism (Section 4.2).

In summary, we make the following contributions:
\begin{itemize}[leftmargin=*]
    \item We introduce a new learning problem named dynamic multimodal continual learning, in which data modalities change across tasks. To the best of our knowledge, this is the first work to investigate this problem and identify its key challenges, particularly spatio-temporal catastrophic forgetting and adaptive multimodal fusion.
    \item We propose a novel method, NeuCME, whose novelty lies in the synergistic integration of modality-combinational rehearsal, multi-gated mixture-of-experts, and task relevance-guided distillation, a combination that endows the model with the ability to address the unique challenges of dynamic multimodal continual learning.
    \item We formulate a new metric to quantify the dynamism of a task sequence. Extensive experiments on four real-world datasets demonstrate the superior performance of the proposed NeuCME over state-of-the-art methods.
\end{itemize}

\begin{figure*}
    \centering
    \includegraphics[width=\linewidth]{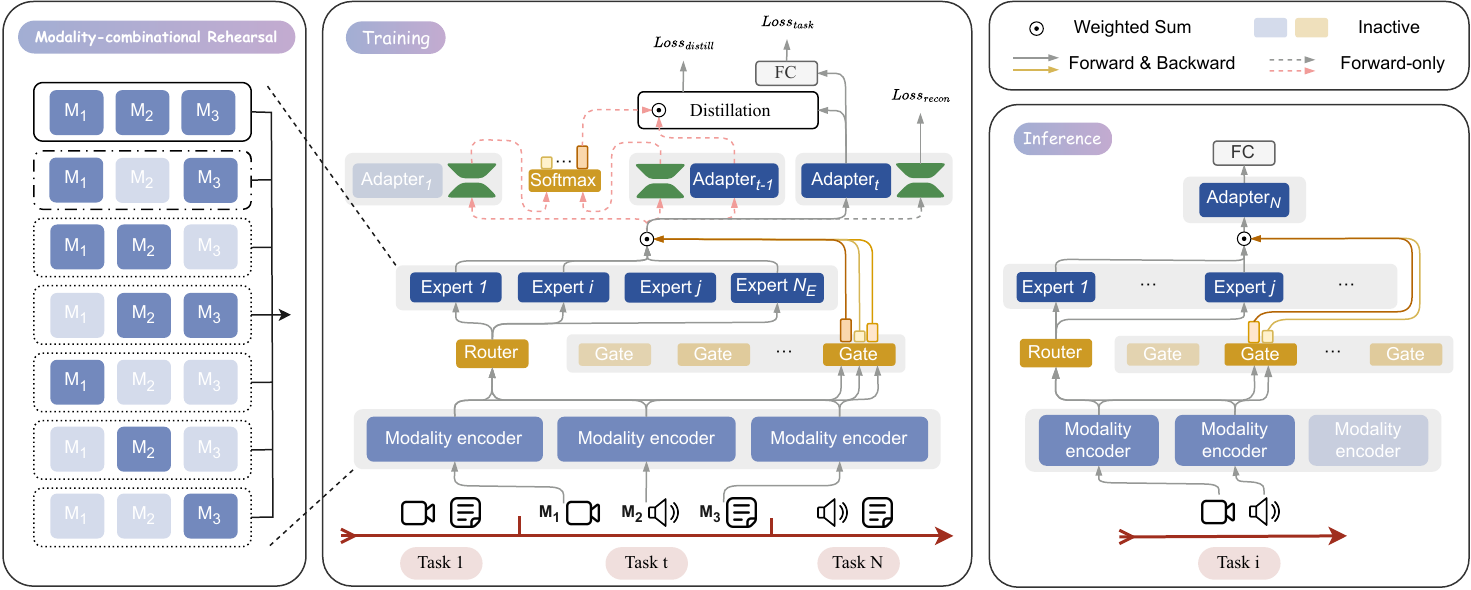}
    \caption{An overview of the proposed NeuCME. It consists of a set of backbone encoders, a \textit{multi-gated Mixture-of-Experts} layer (MMoE), and a \textit{task relevance-guided distillation}. TRAINING (\textit{left}): The whole model is optimized under the proposed \textit{modality-combinational rehearsal}. The joint objective integrates task loss, reconstruction loss, and distillation loss. INFERENCE (\textit{right}): The trained model can handle a variety of tasks with diverse modalities. [Best viewed in color]}
    \label{fig:framework}
\end{figure*}

\section{Related Work}
\subsection{Multimodal Continual Learning}
MCL aims to reduce catastrophic forgetting and promote adaptability by leveraging cross-modal synergies compared to unimodal continual learning~\cite{yu2024recent,sarfraz2025beyond}. Recent progress has largely focused on bimodal continual learning, such as vision-language and video-audio settings~\cite{pian2023audio,chen2023continual,jha2024CLAP}. Among these approaches, data replay-based methods serve as an effective means of addressing catastrophic forgetting~\cite{sarfraz2025beyond,pian2023audio}.
However, existing methods are fundamentally limited to dual-modality settings and therefore cannot generalize to scenarios involving changing modalities. 
In this paper, we bridge this gap by extending MCL to a new learning scenario, namely, dynamic MCL. Concurrently, we propose NeuCME, incorporating a novel modality-level rehearsal to mitigate catastrophic forgetting.

\subsection{Mixture-of-Experts}
MoE typically contains multiple experts and a routing network \cite{shazeer2017outrageously}. It aggregates the expert outputs via a weighted strategy by the routing network. This technique has been increasingly introduced to continual learning. For example, \citet{alujundi2017expert} proposes to train multiple backbones as experts and automatically route test samples to the most relevant one. \citet{chen2023lifelong} utilizes pre-trained experts and gates to preserve previous knowledge. \citet{yu2024boosting} integrates MoE adapters into a pretrained CLIP model. \citet{le2024mixture} explores the integration of MoE with prompt-based continual learning. However, most existing methods are typically tailored for modality-fixed settings. In this paper, we propose a multi-gated MoE, which extends the capability of MoE to handle adaptive multimodal fusion with the novelty of routing first and then multi-gated MoE. 

\subsection{Knowledge Distillation}
Knowledge distillation \cite{hinton2015distilling} is a regularization-based approach to alleviating catastrophic forgetting \cite{li2025conti}. 
Building upon basic preservation methods like LwF \cite{li2017learning} and PODNet \cite{doui2020PODnet}, recent advancements have refined distillation through adaptive feature consolidation \cite{Kang2022afc}, similarity-driven selection \cite{phan2022reminder}, and prompt-based tuning \cite{zhang2025continualdistillationlearningknowledge} to enhance knowledge transfer efficiency.
We propose a task relevance-guided distillation. The proposed distillation strategy selectively utilizes the most relevant adapters from previous tasks as teacher models. This strategy enables positive and effective knowledge transfer while avoiding the injection of negative knowledge from irrelevant historical tasks.

\section{Methodology}


\subsection{Problem Statement} 
\label{sec:problem statement}
Given a sequence of $N$ multimodal tasks $\mathcal{TS}=[T_1, T_2, ..., T_N]$, each task has its own data $T_t=\{D_t, M_t\}$. Here $D_t=\bigcup_{i\in M_t}\{(\mathbf{x}_{i}^t, \mathbf{y}^t)\}$ is the data of the $t$-th task $T_t$, where $\bigcup_{i\in M_t}\{\mathbf{x}_{i}^t\}$ denotes multimodal data, and $\mathbf{y}^t$ is the corresponding label. $M_t \subseteq \mathcal{I}=\{1,2,...,I\}$ denotes the modality set of task $T_t$, where $I$ is the total number of modalities in $\mathcal{TS}$. The $\mathcal{TS}$ is referred to as a dynamic multimodal task sequence if the modalities differ across tasks. For instance, as shown in Figure~\ref{fig:challenges}, the task sequence contains \textit{modality-augmented} subsequence ($M_{t-1}\subsetneq M_t$), e.g., Task 2 vs. Task 1, \textit{modality-reduced} subsequence ($M_{t-1}\supsetneq M_t$), e.g., Task $t$ vs. Task 2, or \textit{modality-mixed} subsequence ($(M_{t-1}\nsubseteq M_t)\land(M_{t-1}\nsupseteq M_t)$), e.g., Task $t$ vs. Task 1. Finally, dynamic MCL aims to learn each task $T_t$ sequentially, adaptively fuse heterogeneous modalities, and effectively accommodate new tasks over time.

\subsection{Framework Overview} Figure \ref{fig:framework} illustrates the overall architecture of the proposed NeuCME, which consists of three key components: (1) \textit{modality-combinational rehearsal} (Section 3.3), which boosts the training of the model, (2) \textit{multi-gated mixture-of-experts} (Section 3.4), which settles adaptive multimodal fusion for heterogeneous modalities, and (3) \textit{task relevance-guided distillation} (Section 3.5), which transfers knowledge from relevant previous tasks to target task. The three components work collaboratively to address the challenges of dynamic MCL. We elaborate on each component below. 

\subsection{Modality-combinational Rehearsal}
\label{sec:combinatorial optimization}

A simple approach to mitigating spatial catastrophic forgetting is parameter isolation across tasks, say allocating a task-specific model to each task $T_t$. This approach reduces the problem to multiple static multimodal continual learning tasks. However, such a method would result in a prohibitive explosion in the number of trainable parameters, usually leading to a space complexity of $O(N)$. Instead of sharing at most $\max\{\vert M_t\vert\}_{t=1}^N$ modality-specific encoders (one per modality) across all tasks, it requires a singleton model for every task, leading to a total of $\sum_{t=1}^N\vert M_t\vert$ encoders, where $\sum_{t=1}^N\vert M_t\vert \gg \max\{\vert M_t\vert\}_{t=1}^N$. 

Our intuition is that modality space across tasks often exhibits subset--superset relationships, i.e., the modalities of some tasks are sub-combinations of others. On the other hand, their backbones can be shared. This gives rise to a novel perspective. Instead of training a model for each task, we can develop a foundation model and train different sub-combinations within it as required per task. Inspired by this observation, we propose a modality-combinational rehearsal to train the model, ensuring robust performance on task $T_t$ and improving its adaptability to varying modality sets.
\begin{equation}
        \arg\min_{\theta}\mathbb{E}_{\mathcal{S}\sim\mathcal{P}(M_t)\setminus\emptyset}[\mathcal{L}(f_{\theta}(\{\mathbf{x}^t_i\}_{i\in\mathcal{S}});\mathbf{y}^t)]
\end{equation}
where $\theta$ is the model parameters and $\mathcal{P}(\cdot)$ is the power set. In practice, at each gradient step, we optimize not only the loss for all modalities of the current task but also the losses for all possible sub-combinations. These losses are then used together to update the whole model $f_\theta$ jointly. Thus, the model would update all $\sum_{t=1}^N\vert M_t\vert$ multi-modal encoders required for different tasks in terms of the $\max\{\vert M_t\vert\}_{t=1}^N$ unimodal encoders, thereby improving computational efficiency.



Compared to parameter isolation, this combinatorial minimization strategy can reduce the number of trainable parameters. Meanwhile, it enables the model to learn a modality-invariant representation space, thereby enhancing the ability of a single backbone to generalize across different modality combinations. Next, we discuss how to integrate multimodal data within a single model.

\subsection{Multi-gated Mixture-of-Experts}
\label{sec:multi-gated MoE}
For the multimodal data $\bigcup_{i\in M_t}\{\mathbf{x}_{i}^t\}$ of the current task $T_t$, we employ a set of extensible modal encoders $\bigcup_{i\in \mathcal{I}}\{\mathcal{F}_i(\cdot)\}$ to derive the representations of the different modalities. 
\begin{equation}
   \bigcup_{i\in M_t}\{\mathbf{r}_{i}^t\} = \bigcup_{i\in M_t}\{\mathcal{F}_i(\mathbf{x}_{i}^t)\} 
\end{equation}
where $\mathbf{r}_i^t \in \mathbb{R}^{d_r}$. To efficiently integrate the multimodal representations of different tasks within a single model, we design a multi-gated Mixture-of-Experts (MMoE) module. The proposed MMoE is configured with $N_E$ distinct experts $\{E_i\}_{i=1}^{N_E}$, a router $\mathcal{R}$ shared by all tasks, and a set of gating networks $\mathbf{G}=\{\mathcal{G}^i\}_{i=1}^{N}$ that can be extended based on the task regarding the set of modalities. The mechanism of MMoE is described below.

For the representation of the current task, we initially utilize the router $\mathcal{R}$ to allocate an expert for the processing of each modal representation. 
\begin{equation}
   \bigcup_{i\in M_t}\{\mathbf{e}_{i}^t\} = \bigcup_{i\in M_t}\{\textcolor{myblue}{E}_{N_i}(\mathbf{r}_{i}^t)\} 
\end{equation}
where $N_{i}=TopOne(\textcolor{myyellow}{\mathcal{R}}(\mathbf{r}_i^t))$ represents the expert index assigned by $\mathcal{R}$.

After obtaining the expert representations $ \bigcup_{i\in M_t}\{\mathbf{e}_{i}^t\}$ for the different modalities, we select the corresponding gating network $\mathcal{G}^t$ based on the current modalities $M_t$ to integrate all expert representations. Thus, the joint representation $\textbf{h}^t$ is derived as follows
\begin{equation}
    \mathbf{h}^t=\sum_{i\in M_t}{{W}^t_{i}}{{\mathbf{e}}_i^t}
\end{equation}
where $W^t = [W^t_i]_{i\in M_t}$ represents the gating weights assigned by $\mathcal{G}^t$, dictating each modal's contribution. The gating weights are then computed as follows
\begin{equation}
    W^t=Softmax(\textcolor{myyellow}{\mathcal{G}^t}(\text{concat}[\mathbf{r}_i^t]_{i\in M_t}))
\end{equation}

As elaborated above, we decouple the functions of the routing network and the gating network. The routing network is responsible for sparsely selecting relevant experts, while the gating network focuses on integrating expert representations of different modalities. Such a method is more advanced than conventional MoE, which cannot simultaneously perform routing and fusion for multiple inputs. 

\subsection{Task Relevance-guided Distillation}
\label{sec:distill adapter}

After obtaining the joint representation $\mathbf{h}^t$ of the task $T_t$, we assign a corresponding adapter to get task-wise representation, i.e.,  $\mathbf{z}^t=\textcolor{myblue}{Adapter^t}(\mathbf{h}^t)$. Finally, we obtain the prediction result through a fully connected layer. The final loss of the task is shown below
\begin{equation}
    \mathcal{L}_{task} = TaskLoss(FC(\mathbf{z}^t), \mathbf{y}^t)
    \label{eq:task-loss}
\end{equation}
To mitigate temporal catastrophic forgetting, we use knowledge distillation to facilitate knowledge forward propagation. Unlike traditional knowledge distillation in the previous model, we select and distill the current adapter from the $k$ most relevant past adapters. Specifically, we assign a task-wise lightweight autoencoder to each $Adapter^t$. Each autoencoder is trained to model the task-wise distribution of $T_t$. This loss function for the autoencoder is defined as
\begin{equation}
    \mathcal{L}_{recon} = re_t=\|\mathbf{h}^t-\mathbf{h}^t_o\|^2
    \label{eq:recon-loss}
\end{equation}
where $\mathbf{h}^t_o=\textcolor{mygreen}{W_{up}^t}\textcolor{mygreen}{W_{down}^t}\mathbf{h}^t$ is the reconstructed feature representation by the autoencoder of the task $T_t$. $W_{up}^t$ and $W_{down}^t$ are the parameters of the autoencoder.

Since each autoencoder is individually learned on the data of the task $T_t$, the resulting reconstruction score $re_t$ reflects the likelihood that the data pertains to the task, with a lower score indicating a higher probability. As a consequence, in performing knowledge distillation, we utilize the reconstruction error of autoencoders to select the $k$ most relevant adapters to serve as the teacher models. The distillation loss is derived as follows
\begin{equation}
    \mathcal{L}_{distill} = MSE(\mathbf{z}^t,\sum_{i=1}^{t-1}W_i^tAdapter^i(\mathbf{h}^t))
    \label{eq:transfer-loss}
\end{equation}

The weights of the previous adapters are determined below
\begin{equation}
    W^t=TopK(\frac{exp(-re_j)}{\sum_{i}^{t-1}exp(-re_i)},k)
    \label{eq:relevant}
\end{equation}
where $k$ is a tunable hyperparameter. By selecting $k$ teachers based on relevance, our distillation strategy allows the current adapter to acquire beneficial knowledge while avoiding the injection of interfering information. During inference, we use the final adapter, which accumulates all learned knowledge. Meanwhile, compared with data replay methods for mitigating forgetting, using lightweight adapters and autoencoders to store and select knowledge is more storage-efficient and avoids privacy concerns.

\myparagraph{Objective Function} Finally, merging Eq.~(\ref{eq:task-loss}), (\ref{eq:recon-loss}), and (\ref{eq:transfer-loss}), we have the joint loss function of our NeuCME as follows
\begin{equation}
    \mathcal{L}_{NeuCME} = \mathcal{L}_{task}+\mathcal{L}_{recon}+\mathcal{L}_{distill} 
    \label{eq:total_loss}
\end{equation}
The pseudo-code for training NeuCME is enclosed in Algorithm 1 of the Appendix. 

\section{Experiments}
\subsection{Experiment Settings}
\label{sec:experi-sets}
We briefly present the experimental settings in this section, with complete details provided in the Appendix.


\subsubsection{Datasets} Since no datasets currently exist for dynamic MCL, we construct a dynamic MCL benchmark based on four publicly available multimodal datasets: CUBICC \cite{palumbo2024deep}, IEMOCAP \cite{busso2008iemocap}, MMEA-CL \cite{10184468}, and CMU-MOSEI \cite{zadeh2018multimodal}. These datasets cover diverse tasks, including image-text understanding, egocentric activity recognition, and multimodal sentiment analysis. Detailed dataset statistics are provided in the Appendix. We use the raw data for CUBICC and MMEA-CL, whereas for IEMOCAP and CMU-MOSEI, we use a pre-trained model to process the videos as executed by \citet{tsai-etal-2019-multimodal} to extract data features. 

\subsubsection{Comparison Methods}
We compare NeuCME against the following SOTA continual learning methods: A-GEM \cite{chaudhry2019efficient}, WA \cite{zhao2020maintaining}, AFEC \cite{wang2021afec}, CLS-ER \cite{arani2022learning}, CMR-MFN \cite{wang2023confusion}, AV-CIL \cite{pian2023audio}, and SAMM \cite{sarfraz2025beyond}. Among them, CMR-MFN, AV-CIL, and SAMM are multimodal continual learning methods based on multi-tower architectures. SAMM serves as a strong baseline, as its design readily accommodates the robust MCL. The remaining methods are representative unimodal continual learning methods. For a fair comparison, we implement these methods using the same multi-tower backbone as NeuCME and fuse representations of different encoders via feature concatenation. For all baseline models, we use the settings as recommended in the original paper for fair comparison.

\subsubsection{Evaluation Metrics} We evaluate all methods with \textbf{Average Accuracy} (AA) and \textbf{Average Forgetting} (AF). AA is the average of the testing accuracy of each task, denoted as $AA=\frac{1}{N}\sum_{t=1}^Na_t$, where $a_t$ is the testing accuracy after training on task $T_t$. AF measures the extent of catastrophic forgetting over previously learned tasks, denoted as $AF=\frac{1}{N-1}\sum_{t=2}^NF_t$, $F_t=\frac{1}{t-1}\sum_{i=1}^{t-1}\mathbf{max}_{\tau\in\{1,....,t-1\}}(a_{\tau, i}-a_{t, i})$, where $a_{\tau, i}$ is the testing accuracy of the $i$-th task after training on the $\tau$-th task.

\subsubsection{Implementation Details} We train all models for 50 epochs with a learning rate of $1 \times 10^{-3}$, employing a 5-epoch warm-up followed by cosine annealing. The batch size is 16 for MMEA-CL and 32 for all other datasets. For our method, the number of experts ($N_E$) in the MMoE module is fixed at 10, and the top-$k$ value in task relevance-guided distillation is set to 3 across all datasets. For the replay-based baselines (A-GEM, CLS-ER, AV-CIL, and SAMM), we adopt an incremental replay buffer strategy, with capacity set to five exemplars per class for all tasks to ensure a fair comparison.

\subsection{Evaluation on Dynamic Task Sequence}
\label{sec:experi-results}
\subsubsection{Dynamism Metric}
To have an in-depth analysis, we formulate an evaluation metric ($\in [0,1]$) to quantify the degree of dynamism in a dynamic multimodal task sequence $\mathcal{TS}$. The metric is defined as follows:
\begin{equation}
\begin{split}
    \mathcal{D}_{metric}(\mathcal{TS})= \frac{1}{N-1}\sum_{i=1}^{N-1}\left(1-\frac{|M_i\cap M_{i+1}|}{|M_i\cup M_{i+1}|}\right)
\end{split}
\end{equation}
The $\mathcal{D}_{metric}$ is derived from the Jaccard distance $d_\mathcal{J}(\cdot, \cdot)$ \cite{Levadow1917dist,Kosub2019A}. It is derived as: $d_\mathcal{J}(M_i, M_{i+1}) =1-\mathcal{J}(M_i, M_{i+1})=1-\frac{|M_i\cap M_{i+1}|}{|M_i\cup M_{i+1}|}$, where $d_\mathcal{J}(M_i, M_{i+1})\in[0,1]$.

Jaccard distance measures the dissimilarity between two sets. In our context, it reflects the degree of modality changes between adjacent multimodal tasks. Then, given a sorted task sequence, we define the degree of dynamism as the average change among all adjacent tasks. In practice, given a target dynamism metric, we generate the corresponding task sequence using a Depth-First Search (DFS) algorithm. The procedure of task sequence generation can be found in Algorithm 2 of the Appendix.

To have a comprehensive evaluation, we divide the dynamism metric into four intervals: (0.00, 0.25), (0.25, 0.50), (0.50, 0.75), and (0.75, 1.00), corresponding to four difficulty levels (denoted as \textbf{L1-L4}) from low to high. We also define an upper bound (=1.0) and a lower bound (=0.0) as extreme cases. We evaluate all methods on this six-level dynamism. The generated dynamic task sequences for each of the six levels are provided in Tables 2, 3, 4, and 5 of the Appendix.

\begin{figure*}[!htb]
    \centering
    \includegraphics[width=\linewidth]{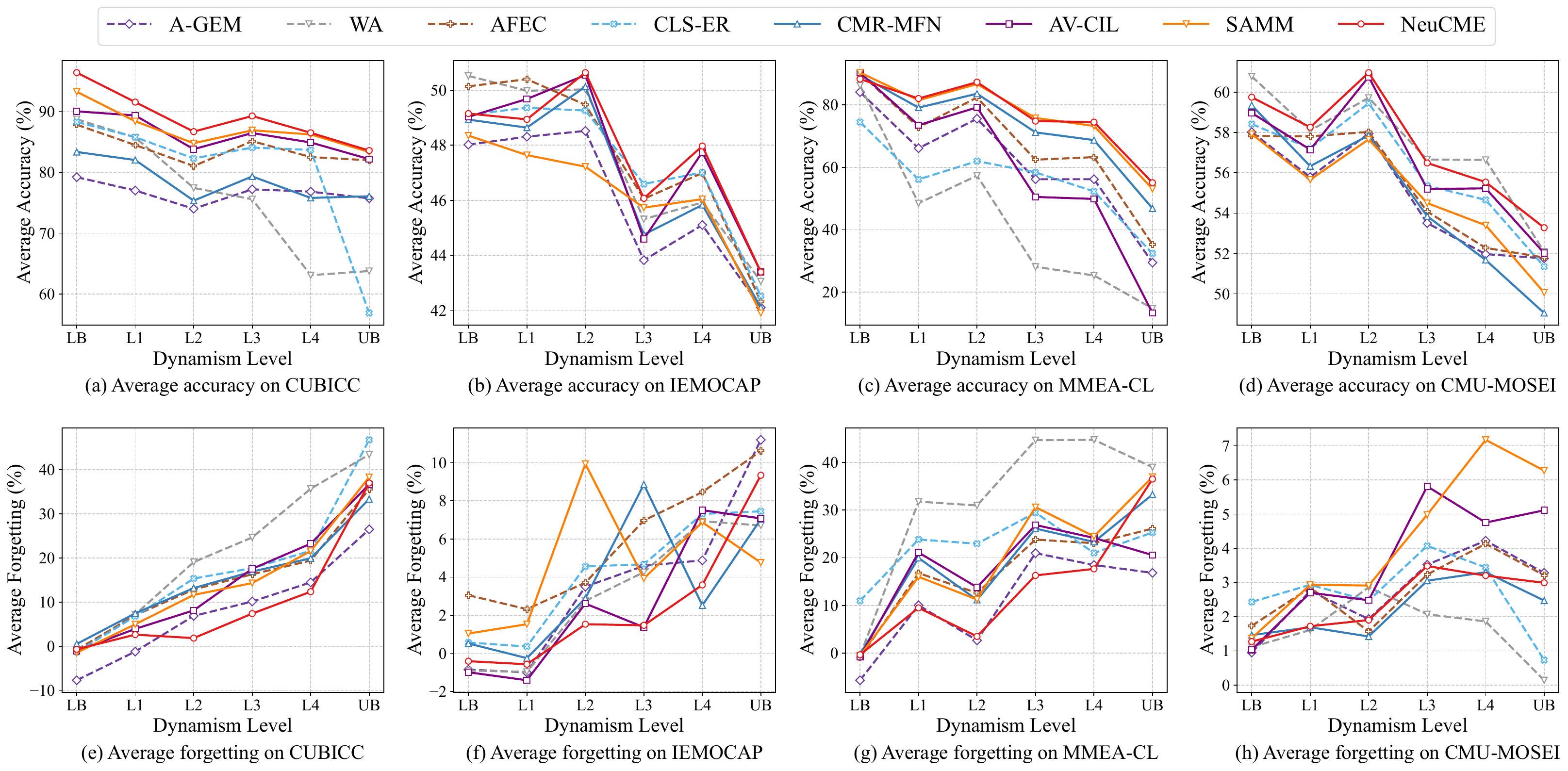}
    \caption{Performance comparison of different methods under varying dynamism levels on four datasets. The \textit{top} row (a-d) shows the average accuracy ($\%$), and the \textit{bottom} row (e-h) shows the average forgetting ($\%$). Different methods correspond to distinct color curves. [Best viewed in color]}
    \label{fig: main results}
\end{figure*}

\subsubsection{Overall Assessment}The evaluation results across different levels of dynamism on each dataset are shown in Figure~\ref{fig: main results}. From the results, we have the following observations:
\begin{itemize}
    \item As the level of dynamism increases, all methods exhibit performance degradation. This substantiates the validity and rationality of the proposed dynamism index. Thus, it provides a reliable metric that can be broadly applied to the future dynamic MCL field.
    \item Most unimodal continual learning methods are inferior to the multimodal counterparts (e.g., NeuCME, SAMM, and AV-CIL), mainly due to the former failing to effectively exploit the complementary information available across multiple modalities.
    \item Our NeuCME outperforms existing MCL methods. The results indicate that NeuCME can effectively tackle adaptive multimodal fusion and mitigate spatio-temporal catastrophic forgetting. 
    \item NeuCME is markedly superior to all baseline methods regarding average accuracy and average forgetting in most cases, particularly in high dynamism (See full results in Tables 6 and 7 of the Appendix).
\end{itemize}


\subsubsection{Analysis of Modality Transitions}
To enable a more fine-grained analysis, we conduct experiments on individual modality transitions. Specifically, for MMEA-CL, the sequences for modality-augmented, modality-reduced, and modality-mixed tasks are \{V\}$\rightarrow$\{V, A\}$\rightarrow$\{V, A, G\}, \{V, A, G\}$\rightarrow$\{V, A\}$\rightarrow$\{V\} and \{A, G\}$\rightarrow$\{V, A\}$\rightarrow$\{V, G\}. For CMU-MOSEI, sequences are \{V\}$\rightarrow$\{V, T\}$\rightarrow$\{V, T, A\}, \{V, T, A\}$\rightarrow$\{V, T\}$\rightarrow$\{V\} and \{V, A\}$\rightarrow$\{V, T\}$\rightarrow$\{T, A\}. The average accuracy results are shown in Table~\ref{tab:single}. From the results, our NeuCME outperforms existing MCL approaches. In particular, our method achieves significantly higher performance on more challenging modality-mixed tasks. This demonstrates that NeuCME learns a modality-robust representation space, effectively resisting negative interference stemming from heterogeneous modality transitions.



\begin{table}
\centering
\begingroup
\small
\setlength{\tabcolsep}{5pt}
\caption{Performance comparison of different methods on various modality transitions. The best and second-best results are marked in \textbf{bold} and \underline{underline}, respectively.}
\label{tab:single}
\begin{tabular}{@{\hspace{5pt}}lcccccc@{\hspace{5pt}}}
\toprule
\multirow{2}{*}{Methods} & \multicolumn{2}{c}{\makecell{Modality-\\augmented}} & \multicolumn{2}{c}{\makecell{Modality-\\reduced}} & \multicolumn{2}{c}{\makecell{Modality-\\mixed}} \\ \cmidrule(lr){2-3} \cmidrule(lr){4-5} \cmidrule(lr){6-7}
          & \makecell{MME\\A-CL} & \makecell{MO\\SEI} & \makecell{MME\\A-CL} & \makecell{MO\\SEI} & \makecell{MME\\A-CL} & \makecell{MO\\SEI}  \\ \midrule
CMR-MFN  & 83.24    & 54.62   & 82.95    & 54.33   & 65.90    & 55.56  \\
AV-CIL  & 82.70    & \underline{55.20}   & 34.86   & \textbf{56.64}  & 56.49  & \underline{56.34}    \\
SAMM  & \textbf{84.28}    & 54.48    & \underline{83.77}    & 56.20   & \underline{69.72}    & 54.84   \\
NeuCME & \underline{83.97} & \textbf{55.48}  & \textbf{84.48}    & \underline{56.35}    & \textbf{70.23} & \textbf{57.71}    \\ 
\bottomrule
\end{tabular}
\endgroup
\end{table}

\subsection{Ablation Studies and Analysis}

In this section, we conduct ablation studies and analyze NeuCME in-depth under the Level-2 dynamism setting.
\begin{figure*}
    \centering

    \begin{minipage}{0.49\textwidth}
        \centering
        \includegraphics[width=\linewidth]{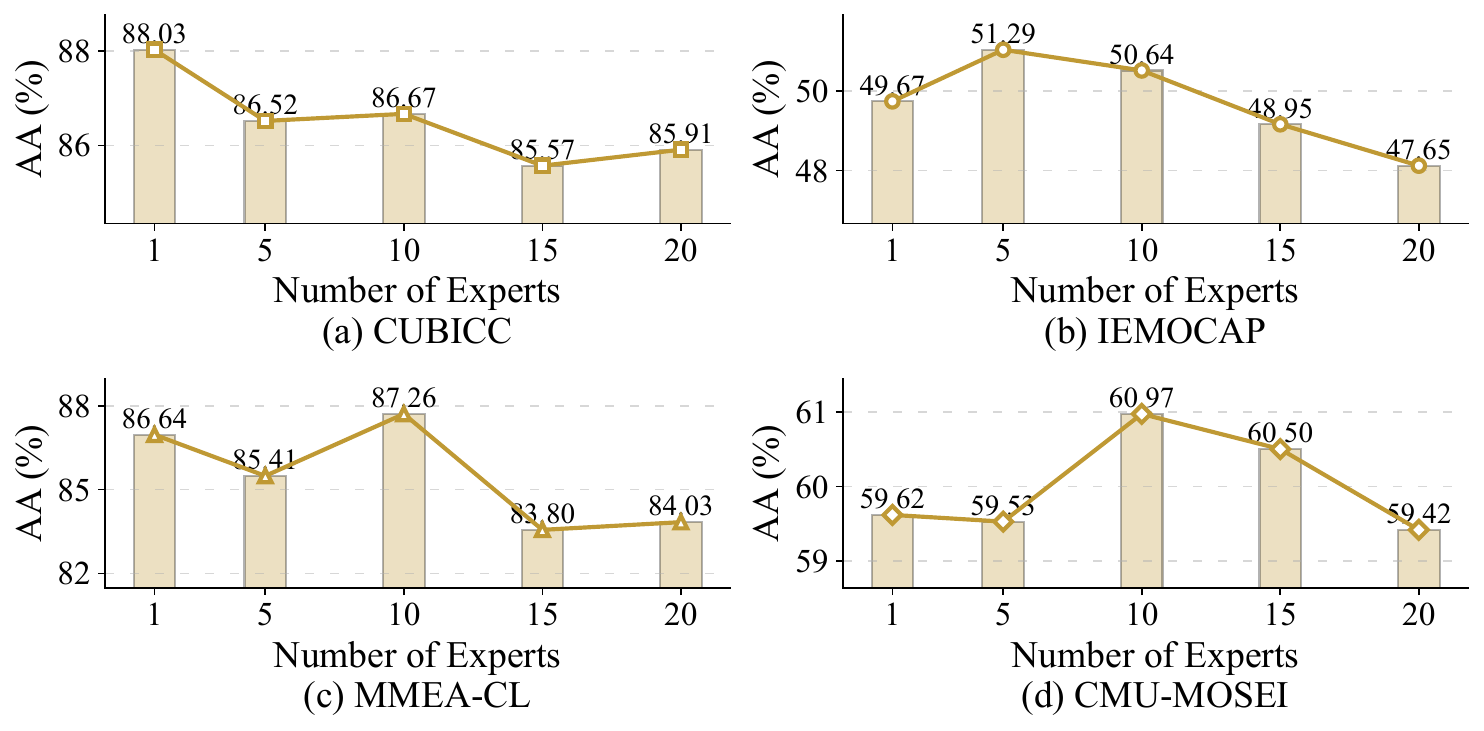}
        \caption{Analysis of the number of experts.}
        \label{fig:experts analysis}
    \end{minipage}
    \hfill
    \begin{minipage}{0.49\textwidth}
        \centering
        \includegraphics[width=\linewidth]{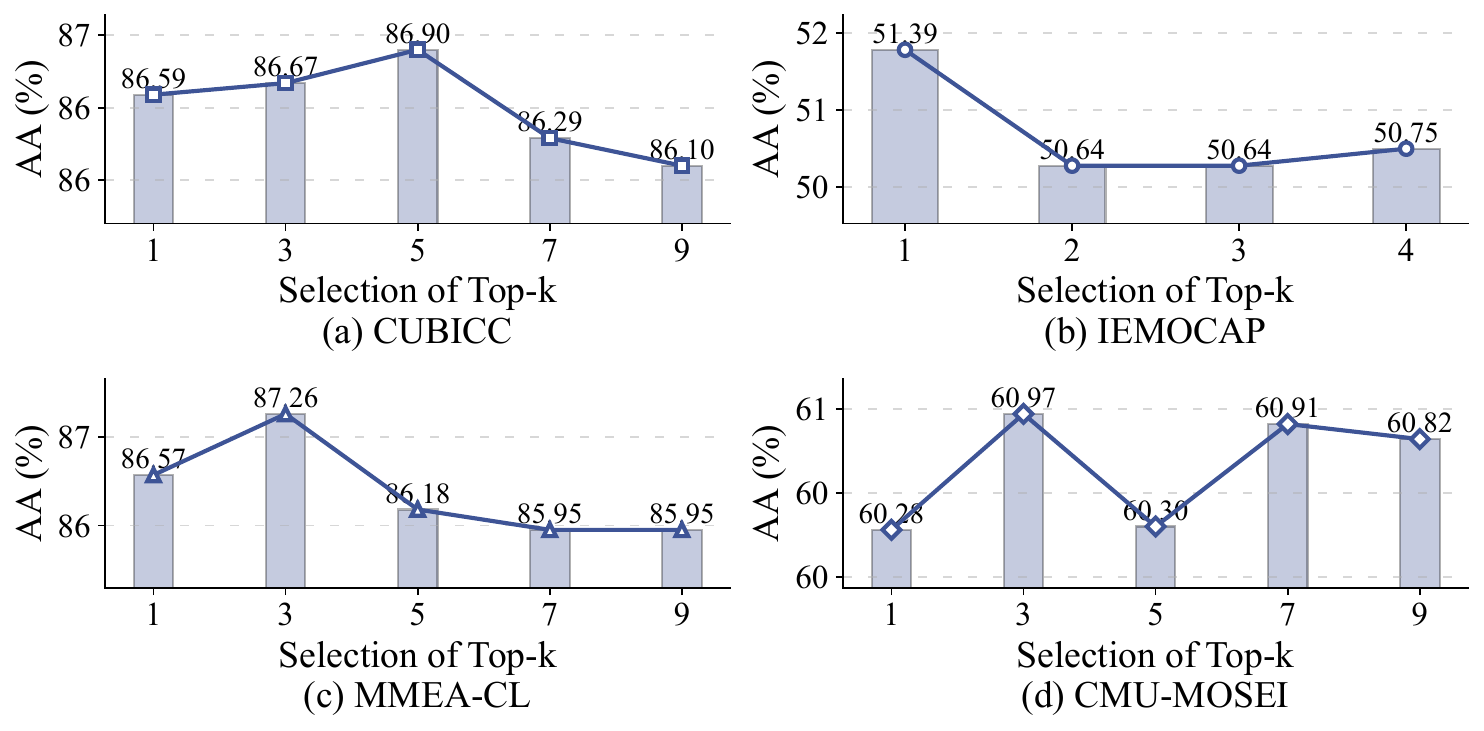}
        \caption{Analysis of the selection of top-$k$.}
        \label{fig:topk analysis}
    \end{minipage}

\end{figure*}

\subsubsection{Ablation Study} 
\begin{table}
\centering
\small
\setlength{\tabcolsep}{5pt}
\caption{Ablation study on the variants of NeuCME under Level-2 dynamism. Best results are in \textbf{Bold} fonts.}
\label{tab:ablation}
\begin{tabular}{@{\hspace{5pt}}lcccc@{\hspace{5pt}}}
\toprule
Variants & {CUBICC} & IEMOCAP & MMEA-CL & MOSEI  \\ \midrule
w/o MCR  & 84.89    & 48.00  & 78.80  & 59.08  \\
w/o MMoE  & 86.59    & 47.34  & 87.03  & 58.86  \\
w/o TRD  & 86.25    & 49.27  & 86.72  & 59.79  \\
NeuCME & \textbf{86.67} & \textbf{50.64} & \textbf{87.26} & \textbf{60.97}  \\ 
\bottomrule
\end{tabular}
\end{table}

We conduct an ablation study to measure the performance gain by each of the three key components: modality-combinational rehearsal (MCR), multi-gated Mixture-of-Experts (MMoE), and task relevance-guided distillation (TRD). We remove each component from NeuCME and then evaluate the performance of each resulting variant on four datasets using the average accuracy. The evaluation results are presented in Table \ref{tab:ablation}. From the results, we observe that removing any component leads to a clear performance degradation, confirming that every component plays a critical role in the whole model. Specifically, MCR demonstrates the most significant performance on raw data (i.e., CUBICC and MMEA-CL), closely followed by TRD. For feature data (i.e., IEMMCAP and MOSEI), the MMoE proves to be the most effective, with MCR achieving the second-best gains.

\subsubsection{Analysis of Experts Number} To investigate the impact of the number of experts ($N_E$) in MMoE, we fix the top-$k$ value at 3 and then conduct a grid search over $N_E\in$ \{1, 5, 10, 15, 20\}. As shown in Figure \ref{fig:experts analysis}, we observe that an excessive number of experts (beyond 10) does not yield additional performance improvement. Excessive expert count often introduces redundancy, leading to expert imbalance and degrading returns despite increased model capacity. Besides, the results show a correlation between $N_E$ and the number of modalities. Specifically, a smaller number of experts is sufficient for tasks with fewer modalities (e.g., CUBICC). Whereas for tasks involving more modalities, a moderate number of experts (e.g., $N_E$ = 10) is optimal.


\subsubsection{Selection of Top-$k$ Value} We investigate the effect of the selection of the top-$k$ value in TRD by conducting a grid search of $k \in $ \{1, 3, 5, 7, 9\}, except \{1, 2, 3, 4\} for the IEMOCAP dataset (as the size of this dataset is small). The results are shown in Figure \ref{fig:topk analysis}. Our analysis reveals a clear dependency on task sequence length. Specifically, we observe that $k$ = 1 is most appropriate for short-sequence tasks (e.g., IEMOCAP), while $k$ = 3 is better suited for long-sequence tasks. Furthermore, an overly large $k$ value leads to a performance degradation. An excessively large $k$ value leads to the selection of irrelevant task adapters, thereby risking the injection of interfering knowledge. 

\subsubsection{Analysis of Gating and Distillation}

\begin{table}
\centering
\small
\caption{Analysis of gating and distillation study.}
\label{tab: gating and distillation}
\begin{tabular}{@{\hspace{5pt}}llcccc@{\hspace{5pt}}}
\toprule
\multicolumn{2}{c}{Settings}  & MMEA-CL & CMU-MOSEI  \\ \midrule
\multirow{2}{*}{Gate type} & single-gated   & 86.65  & 58.17  \\
& multi-gated & \textbf{87.26} & \textbf{60.97}   \\ \midrule
\multirow{4}{*}{Distillation} & all   & 85.95  & 60.82 \\
& prior  & 84.72  & 60.37 \\
& random  & 85.72  & 59.79 \\
 & top-k  & \textbf{87.26} &  \textbf{60.97}  \\ 
\bottomrule
\end{tabular}
\end{table}

We conduct an analysis study to evaluate the impact of different gating mechanisms and distillation strategies. For the gating mechanisms in MMoE, we compare the proposed multi-gate with a single-gate. For distillation strategies in TRD, we compare our top-$k$ selection with all selection, prior selection, and random selection. The average accuracy results are shown in Table \ref{tab: gating and distillation}. Our proposed multi-gated MoE demonstrates its advantage in adaptive multimodal fusion. Meanwhile, our top-$k$ selection outperforms the widely used strategies and highlights the benefits of selecting relevant tasks for knowledge transfer.


\subsubsection{Efficiency Analysis}
We further evaluate the computational efficiency of different methods by comparing the training time per epoch (s) and the number of trainable parameters (M). The experimental results are summarized in Table \ref{tab: efficiency}. Although NeuCME introduces modality-combinational rehearsal and progressively expands lightweight experts during continual learning, it consistently achieves superior performance with lower training time under comparable parameter budgets. These results demonstrate that NeuCME achieves an effective trade-off between computational efficiency and model flexibility, outperforming methods based on attention-intensive architectures (e.g., AV-CIL and CMR-MFN) or more complex training pipelines (e.g., SAMM).


\begin{table}
\centering
\small
\caption{Runtime and parameters of different methods under Level-2 dynamism. Best results are in \textbf{Bold} fonts.}
\label{tab: efficiency}
\begin{tabular}{@{\hspace{5pt}}lcccc@{\hspace{5pt}}}
\toprule
\multirow{2}{*}{Methods} & \multicolumn{2}{c}{CUBICC}  & \multicolumn{2}{c}{MMEA-CL}   \\ \cmidrule(lr){2-3} \cmidrule(lr){4-5}  
& Runtime & Parameter & Runtime & Parameter \\ \midrule
CMR-MFN  & 129.86 & 23.97     & 156.52 & \textbf{20.32}    \\
AV-CIL  & 153.31 & \textbf{21.52}    &  180.30 & 23.75   \\
SAMM  &  178.94 & 24.44      &  211.57 & 22.92    \\
NeuCME & \textbf{48.41} & 27.04  & \textbf{134.71} & 23.45 \\ 
\bottomrule
\end{tabular}
\end{table}



\section{Conclusion}
This paper investigated a new yet nontrivial learning problem called dynamic multimodal continual learning and revealed its key challenges, particularly spatio-temporal catastrophic forgetting and adaptive multimodal fusion. To address this problem and the challenges, we proposed a novel method NeuCME, whose novelties are encapsulated in its amalgamation of modality-combinational rehearsal, multi-gated mixture-of-experts, and task relevance-guided distillation. Experimental results on real-world datasets show the effectiveness of NeuCME and demonstrate its capability to handle dynamic multimodal continual learning scenarios. 


\bibliography{aaai2027}


\end{document}